# MULTIFORM ADAPTIVE ROBOT SKILL LEARNING FROM HUMANS

**Leidi Zhao, Raheem Lawhorn, Siddharth Patil**
**Steve Susanibar, Lu Lu, Cong Wang**

ECE and MIE Departments
New Jersey Institute of Technology
Newark, NJ 07102
Email: {lz328, rsl3, sp899, ss976, lulu, wangcong}@njit.edu

**Bo Ouyang**

College of Electrical and Information Engineering
Hunan University, Changsha
Hunan Province, 410082 China
ouyangbo@hnu.edu.cn

## ABSTRACT

*Object manipulation is a basic element in everyday human lives. Robotic manipulation has progressed from maneuvering single-rigid-body objects with firm grasping to maneuvering soft objects and handling contact-rich actions. Meanwhile, technologies such as robot learning from demonstration have enabled humans to intuitively train robots. This paper discusses a new level of robotic learning-based manipulation. In contrast to the single form of learning from demonstration, we propose a multiform learning approach that integrates additional forms of skill acquisition, including adaptive learning from definition and evaluation. Moreover, going beyond state-of-the-art technologies of handling purely rigid or soft objects in a pseudo-static manner, our work allows robots to learn to handle partly rigid partly soft objects with time-critical skills and sophisticated contact control. Such capability of robotic manipulation offers a variety of new possibilities in human-robot interaction.*

## INTRODUCTION

The capability of learning new skills from humans is of fundamental necessity to robots in ubiquitous human-robot interactive coexistence. Without the need for extensive coding, robots should be able to acquire manipulation skills from humans to (1) handle objects that are more complex than just a single rigid body or a single piece of soft material, (2) utilize sophisticated contact control and timing to manipulate objects, going beyond simple firm grasping, and (3) fuse complementary data from different sources and in different forms to complete the knowledge. So far, these expectations have not been fully achieved.

During the early history of robotic manipulation, robots were used to handle single rigid-body objects with firm grasping. A recent development is the capability of handling soft and deformable objects such as clothes and ropes [1, 2, 3], while another pursuit is the utilization of contacts, including colliding [4, 5] and sliding [6] to manipulate objects in subtle manners. Meanwhile, in terms of learning from humans, a major pursuit has been in enabling natural teaching, in which humans teach robots in intuitive manners similar to human-human teaching without extensive coding. Specifically, most research have been focused on the concept of robot learning from (human) demonstration (a.k.a. apprenticeship learning and imitation learning [7]). Additionally, the use of skill learning is emphasized, in which a learning agent is expected to extract generic strategies from demonstrations instead of simply recording and replaying the demonstrated actions, so as to handle variations such as different layouts of the rope in a tying task [8, 9].

In terms of achieving sustained ubiquitous human-robot coexistence, the state-of-the-art developments have several limitations: (1) Learning only from single source and in a single form: Learning from demonstration is usually the only form of skill acquisition. This single learning form alone is insufficient for human learners let alone for robot learners, which have very limited perception intelligence. In addition to demonstration, human learners usually also need abstract explanations in verbal or symbolic forms, as well as evaluation of their practice to effectively acquire new skills [10]. The same framework should be applied to robot learning.

(2) Handling only single bodies that are purely rigid or soft: Despite the recent developments of robot manipulation in the handling of soft objects such as clothes and rope, robots have not yet learned from humans to manipulate gadgets that consist of both rigid and soft parts. As shown in Fig. 1, such objects are very common in human lives.

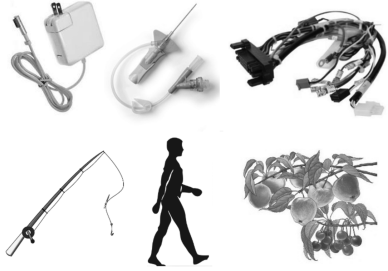

**FIGURE 1.** OBJECTS THAT ARE PARTLY RIGID AND PARTLY SOFT, INCLUDING HUMAN BODIES, ANIMALS, AND PLANTS.

(3) Limited capability on learning highly dynamic skills: In most works so far, human mentors can only teach robots skills that do not require much dynamics and/or critical timing. Usually the robot can only carry out the learned skill at such a slow speed that demo videos of its operation need to be played back at ×10 or higher speed. Meanwhile, the few works that did achieve highly dynamic object maneuvering such as ball-pitching [11], throwing and re-grasping [12], and flipping with fingers [13] are heavily hard-coded and rely on high-fidelity ad-hoc models. Such approaches cannot efficiently facilitate ubiquitous robot learning from humans.

In regard to the aforementioned issues in robot learning, we proposed a multiform robot learning scheme [14] that goes beyond the single form of learning from demonstration and introduces additional forms of learning, including learning from definition and learning from evaluation. These forms of learning complement each other and allow humans to intuitively teach robots in a manner similar to human-human teaching [10]. In addition, adaptive techniques are used to enable autonomous revision of the original definition provided by the human mentor. The proposed strategy helps robots to master complex skills that they have difficulties acquiring thus far. Specifically, we aim for highly dynamic and contact-rich skills and the handling of objects with significantly inconsistent stiffness such as tethered tools, human bodies/animals/plants, and even martial arts instruments. This paper is continued from an earlier introduction of our work in [14]. In particular, the formulation of adaptive Petri nets (APN) for adaptive correction of human definition is further developed. Additional test designs and results are presented and discussed.

## MULTIFORM ROBOT LEARNING FROM HUMANS

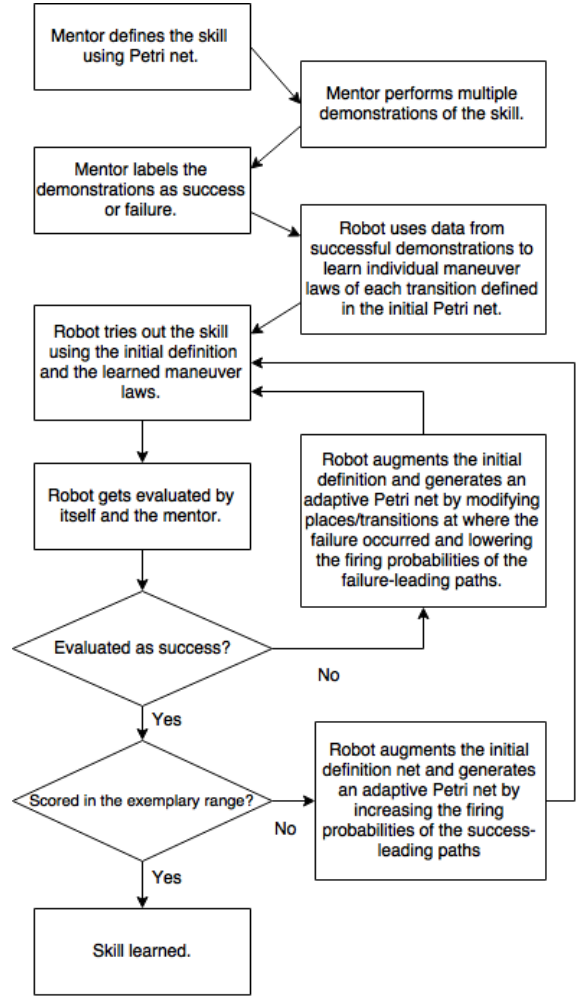

**FIGURE 2.** MULTIFORM ROBOT LEARNING FROM HUMANS

### Adaptive Learning from Definition

When humans teach new tasks and skills to each other, the mentor usually divides the task into multiple sub-procedures that each feature relatively simple patterns. Then the mentor teaches the learner about the interconnection of the sub-procedures, as well as the details of each sub-procedure. Every sub-procedure can be specified using a set of states, connecting transitions, and conditions. The completion of a sub-procedure triggers the subsequent action(s). Such a learning framework describes a task or skill as a discrete event dynamic system with each transition in the system governed by a continuous maneuver strategy. Many tools are available to model discrete event dynamic systems, including state-transition diagrams, reactive flow diagrams, and Petri nets. In particular, Petri nets are abstract enough to be composed intuitively by humans while also sufficiently symbolic for machine algorithms to parse. In this work, we use Petri nets to facilitate the human definition to robots.

Despite that standard Petri nets have shown great efficiency in modeling robotic tasks (e.g. [15], [16]), they have not been used to teach robots highly dynamic skills such as handling rigid-soft objects with contact-rich and time-critical maneuvers. Standard Petri nets also lack the flexibility for modeling a system with uncertainties. During the learning process, the robot might reach incorrect states which cannot be corrected simply through repeating the procedure. The initial definition that human mentors specified may lack necessary steps or contain incorrect or superfluous sub-procedures. In this case, the robot would always end in failure if the definition of the skill/task is not modified. In regard to this issue, we propose an adaptive Petri net (APN) to allow the robot to modify the initial definition according to its experience and through perceiving the human mentor's intention in the demonstrations.

A standard Petri net is typically defined as a 4-tuple: $M = (P, T, A, m_0)$, where $P$ is the set of places, $T = \{t_1, t_2, \ldots, t_n\}$ is the set of transitions, $A$ is the incident matrix specifying the relationship among places and transitions, and $m_0$ is the initial marking. In order to facilitate adaptation, we propose an adaptive Petri net (APN) with additional variable sets that allow adaptive modification. In [14], we introduced a 5-tuple APN with adjustable variables controlling firing probabilities. Here, we extend the idea and introduce a 6-tuple APN: $M = (P, T, A, m_0, \Lambda, C)$, where $\Lambda$ is the set of firing probability $\lambda$'s of the transitions, and $C$ is the set of conditions for each transition. The places in the APN are specified by the state variables of the robot and the object such as the layout, velocity, position, and so on. Each transition in the APN features a relatively coherent motion pattern and can be realized using a single motion/force control law. The firing probability $\lambda$ of each transition determines the chance that the transition is executed in the trial practice.

Figure 3 explains the proposed APN. $P_0$, $P_{F(\text{success})}$, and $P_{F(\text{fail})}$ denote the initial, final successful, and final failure places respectively. The shaded areas are the unknown places and transitions that will be augmented. Initially, the human mentor defines the skill to be learned by the robot by segmenting the task into sub-procedures and constructing the initial Petri net. Following the initial Petri net, the robot attempts a trial of the skill from $P_0$. When the state variables in a place satisfy the necessary condition, the robot advances to make a stochastic decision on which transition, as a function of $\lambda$ should be fired. The robot continues with this procedure until a final place is reached. Trials concluding in $P_{F(\text{fail})}$ indicate robot maneuvers that are incorrect, and thus where augmentation is required. The robot then creates an APN from the initial Petri net by autonomously modifying the transitions or places in the Petri net, and/or changing $\Lambda$ and $C$ based on the data from the demonstrations. The transitions in which errors occurred will have their corresponding $\lambda$'s reduced (while others increased), lowering the chances of these transitions being fired in future trials. This continues until the trials reach a success. A new place can also be added in the Petri net if a new transition is added in a part that all places have no relationship.

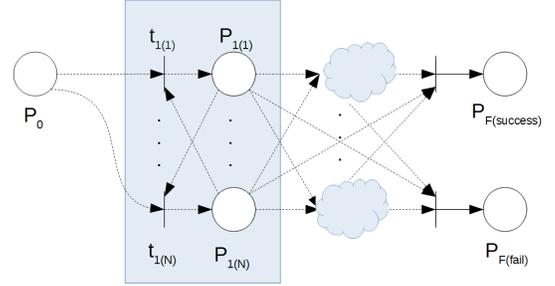

**FIGURE 3**. ADAPTIVE PETRI NET

A related idea is introduced in [17] regarding robot error recovery in manipulation tasks defined by Petri nets that have implicit transitions constantly resulting in failure. In the work, the robot prompts the human to specify additional transitions so that the process could be brought back to known places in the Petri net. The approach and our proposed method both allow new places and transitions to be added to the originally defined Petri net. However, [17] requires a human to manually edit the Petri net while our approach aims for autonomous correction without the need of human intervention.

**Nonparametric Learning from Demonstration**

The goal of learning from demonstration is to extract control laws from the human demonstration data. Each transition specified in the Petri net definition is supposed to have a relatively coherent motion pattern so that it can be governed by one single control law. In order to avoid the limit of model-based and structured control and acquire versatility, we propose to use nonparametric learning methods. In particular, Gaussian process regression (GPR) [18] is used. The demonstrations from the human mentor generate a set of training data $\{(x_i, y_i) : i = 1, 2, \ldots, n\}$, where each $x_i$ consists of the state variables (mostly motion variables of the object being manipulated) and $y_i$ consists of corresponding maneuvering control variables (applied by the human). After training the hyperparameters of GPR, it can be used to determine the proper control signals to realize a desired motion.

The squared exponential kernel function

$$k(x_i, x_j | \theta) = \sigma_f^2 \exp\left(\frac{-(x_i - x_j)^2}{2l^2}\right) \quad (1)$$

is used, where $x_i$ and $x_j$ are from two data points. $\theta = \{\sigma_f, l\}$ includes the hyperparameters to be trained, where $\sigma_f$ is the allowable covariance and the $l$ is a distance parameter. The covariance matrix among the training data points can be built as

$$K = \begin{bmatrix} k(x_1, x_1) & \cdots & k(x_1, x_n) \\ \vdots & \ddots & \vdots \\ k(x_n, x_1) & \cdots & k(x_n, x_n) \end{bmatrix} \quad (2)$$

For a given $x_*$, the values of $y_*$ to be predicted follow a joint Gaussian distribution (with the training data) as

$$\begin{bmatrix} y \\ y_* \end{bmatrix} \sim \mathcal{N}\left(0, \begin{bmatrix} K & K_*^\mathsf{T} \\ K_* & K_{**} \end{bmatrix}\right) \quad (3)$$

where $y = [y_1, y_2, \ldots, y_n]^\mathsf{T}$. The conditional probability $p(y_*|y)$ following the joint distribution is

$$p(y_*|y) \sim \mathcal{N}(\mathbb{E}[y_*], \mathrm{var}(y_*)) \quad (4)$$

where

$$\mathbb{E}[y_*] = K_* K^{-1} y \quad (5)$$
$$\mathrm{var}(y_*) = K_{**} - K_* K^{-1} K_*^\mathsf{T} \quad (6)$$

are the mean (predicted value) and variance (uncertainty) of $y_*$ respectively. $K_{**} := k(x_*, x_*)$ and $K_* := [k(x_*, x_1), \ldots, k(x_*, x_N)]$.

A particular issue in robot learning from human demonstration is the correspondence problem [19]. A direct mapping of human motion to robot motion is usually not possible due to the differences on kinematic structures of the two. We avoid this problem by separating the motion of the robot's end-effector from that of the actuating joints. Only the maneuvering skill (motion and force) of the human's hand is learned and replicated by the robot end-effector, while the motor actuation required to perform the learned maneuver is learned by a separate auxiliary learning action that utilizes the experience motion data of the robot itself rather than the human mentor's demonstration. Such an auxiliary learning action is explained in Fig. 4.

**Learning from Evaluation**

Similar to humans learning from humans, a third measure in our proposed approach is robot learning from human and self-evaluation. After each demonstration, the human mentor first labels each sub-procedure in his/her own demonstration as success or failure, without specifying any explicit judging criteria. The robot only learns from the successful demonstration to extract the control laws. Meanwhile, the robot also examines the failing demonstrations and learns about the judging criteria, which is then used by the robot to evaluate its own trial practice. This allows the robot to clean its own experience data and improve, so as to avoid future failures.

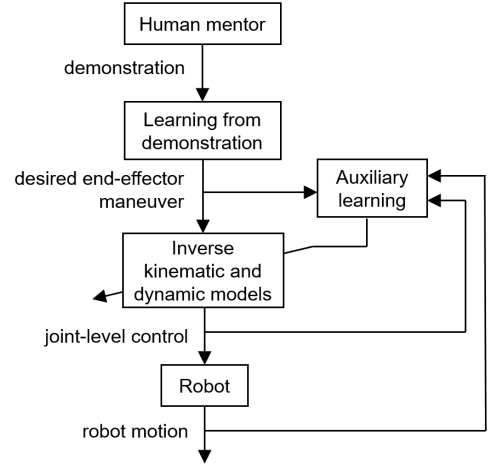

**FIGURE 4**. THE AUXILIARY LEARNING ACTION

Clustering maps are generated as the perceived criteria of success and failure. Once trials are conducted by the robot, the performance will be graded by the robot itself as well as by the human mentor. The new labels from the human mentor will then be used to incrementally revise the learned qualification rules. A support vector machine (SVM) is used by the robot to learn the judging criteria from human-labeled demonstrations. A pool of different learning kernels for evaluating the relevance is provided, from which the learning agent selects to achieve the best clustering result for different motion patterns. In particular, the radial basis function kernel [20] has been used often. A radial basis function kernel has the form $\exp(-\gamma||x_i - x_j||^2)$, where $\gamma$ is the hyperparameter that controls the kernel's width, and $x_i$ and $x_j$ are from two data points.

**VALIDATION TESTS**

Two tests have been conducted to validate the proposed robot learning method, including (1) a human teaching an actual robot manipulator to swing up and balance an inverted pendulum, and (2) a human teaching a simulated robot manipulator to flip a nunchaku, notable for its soft-rigid structure requiring highly dynamic and contact-rich skills for handling.

**Inverted Pendulum Swing-up and Balancing**

Due to its dynamic nature, the swing-up and balancing of an inverted pendulum has been a popular test for robot learning strategies. As shown in Fig. 6, a 6-axis AUBO i5 robot manipulator is used to carry an inverted pendulum. The robot provides an open-architecture control interface driven by a Controller Area Network (CAN) bus, allowing torque, velocity, or position level control. A National Instruments CAN PCI interface and MAT-LAB/Simulink Vehicle Network Toolbox are used to facilitate a CAN-based real-time control system at a sampling rate of 1 kHz.

Figure 5 shows the Petri net definition of the skill, which consists of the swing-up and balancing phases. $P_0$ denotes the initial state, $P_1$ and $P_2$ represent the swung-up and balanced states (with success/failure judging conditions) respectively. $P_{F(success)}$ and $P_{F(failure)}$ are the final success or failure status. The state variables in the places are the angle $\theta$ of the pendulum and its angular velocity $\omega$. $t_0$ denotes the starting move. $t_1$ is the switching action from swing-up to balancing. $t_2$ and $t_3$ are the repeating swing-up and balancing moves respectively. $t_4$ or $t_5$ fires when the swing-up state or the balanced conditions cannot be reached over time and lead to the $P_{F(failure)}$ state. The successful stop action $t_6$ fires when the balanced state is maintained steadily for a certain amount of time. Transitions $t_4$, $t_5$ and $t_6$ all lead to the end of the maneuver.

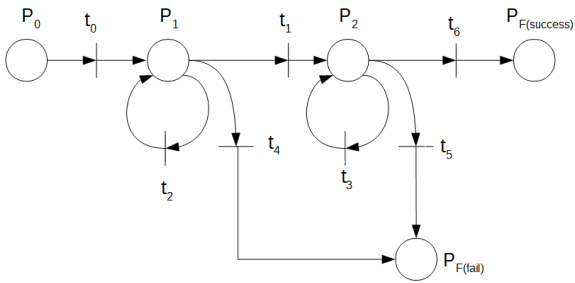

**FIGURE 5**. THE SKILL OF SWING-UP AND BALANCING OF AN INVERTED PENDULUM DEFINED BY A HUMAN MENTOR

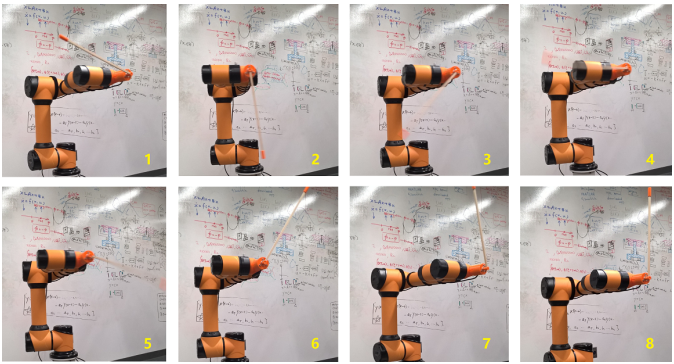

**FIGURE 6**. A ROBOT LEARNED TO SWING UP AND BALANCE AN INVERTED PENDULUM

Multiple demonstrations of the skill are performed by a human mentor using joystick control. The human mentor labels the individual performances in his/her own demonstrations as success or failure. The control laws of each transition specified in the Petri net are trained using data from the successful demonstrations. In addition, the learning agent learns the judging criteria from the labeled data and allows the robot to grade its own performance. Starting from the initial Petri net definition and demonstration data, the robot attempts repeated trials of the skill. Each time a trial is completed, the robot grades its own performance using the learned criteria.

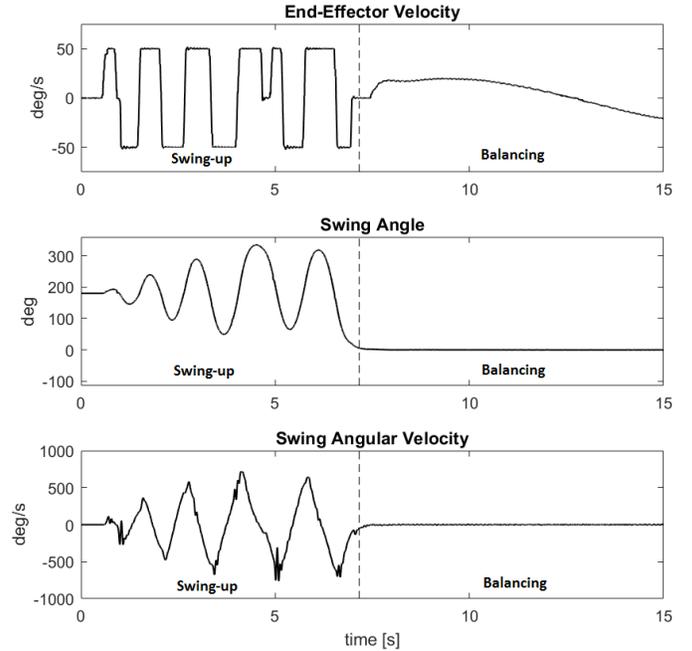

**FIGURE 7**. MOTION VARIABLES OF THE ROBOT AND THE PENDULUM

Depending on the trial results, the robot may modify the initial Petri net definition (Fig. 5) provided by the human mentor using the APN method introduced earlier. As an example, in the case that the human mentor defined the condition of firing $t_1$ (swtiching to balancing) to require exact upright position and absolute zero angle velocity of the pendulum (i.e. a perfect swing-up), the system would almost never get the chance to proceed to $P_2$, and the process would end up failing (after repeating $t_2$ a certain number of times) in every trial. Upon the detection of constant failure, the robot starts to modify the original definition using the proposed APN. First, the learning agent identifies that transition $t_1$ has never been fired, the condition to fire it has never been satisfied, and place $P_2$ has never been visited. As $t_1$ is already defined as the only transition connecting $P_1$ and $P_2$ (as specified in $A$), no new transition is added, and adjustment of $\Lambda$ would induce no effects. That leaves the learning agent to adjust $C$ by examining the data collected from human demonstration. In particular, an SVM is used to learn a new condition (specified by the state variables) by regression from the demonstrated switching action corresponding to $t_1$. The resulting APN features a feasible definition of $C$ and allows successful triggering of $t_1$. Figure 6 shows a successful performance carried out by the robot after the learning. Figure 7 shows the motion variables of the robot and the pendulum, in which the transition from swing-up to balancing can be clearly seen.

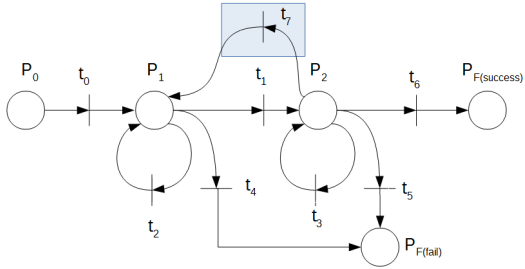

**FIGURE 8.** REVISED PETRI NET

Sometimes the process is also affected by excessive external disturbance and forced out of a known sub-procedure. In particular, before $t_6$ is fired (upon steady balancing over a certain amount of time), the balanced status can be broken due to external disturbance, which might be too intense and the status cannot be recovered by firing $t_3$. Such a situation is not specified in the initial definition. In order to recover, the learning agent first identifies $P_2$ in the Petri net as where the issue occurs. The state variables after the incident occurs are examined. Specifically, their correlation to each place in the Petri net is evaluated, which indicates that the status belongs to $P_1$. The learning agent then adds a new transition $t_7$ from $P_2$ to $P_1$ as shown in Fig. 8, which recovers the process by backing up to the swing-up phase.

**The Nunchaku Challenge**

Compared to the basic inverted pendulum test, the second test is a significantly more challenging test that distinguishes the strength of the proposed method from conventional single form robot learning solely based on human demonstrations. The test features the handling of a partly rigid partly soft gadget, and the use of highly dynamic contact-rich maneuver. A simulation study is carried out in which a robot learns the nunchaku flipping skill from a human mentor. The nunchaku is a traditional Okinawan martial arts weapon widely known due to its depiction in film and pop culture. It consists of two rigid rods and a soft rope (or chain) connecting the rods. Nunchaku flipping (Fig. 9) is a highly dynamic maneuver that requires sophisticated contact control and accurate timing. The skill is quite difficult even for human learners. Acquiring such a skill is an extreme challenge to robotic learning and manipulation capability.

The initial Petri net definition for nunchaku flipping is defined as shown in Fig. 10. The skill is divided into three major sub-procedures: (1) swing-up, (2) chain rolling, and (3) regrasping. $P_0$ represents the initial state of the robot's end-effector holding one of the rods. The swing-up is defined by the starting move $t_0$, the swinging move $t_1$, and the judging conditions in $P_1$. The chain rolling is defined by the releasing action $t_2$, the back palm contact control $t_3$, and the judging conditions in $P_2$. The regrasping is defined by the regrasping action $t_4$, the final successful stopping action $t_5$, and the judging conditions in $P_3$. Transitions $t_6$, $t_7$, $t_8$ lead to the failed end after certain counts of unsuccessful local repetition. $P_{F(success)}$ and $P_{F(failure)}$ are the final success and failure states. Note that this definition shares a similar structure with the inverted pendulum example. Such a common structure makes composing Petri net definitions for various skills relatively easy to humans.

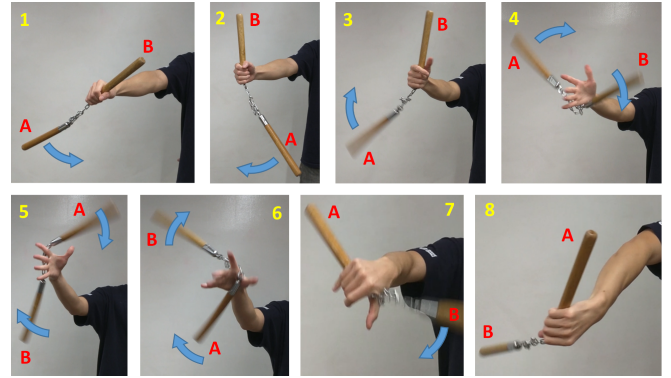

**FIGURE 9.** NUNCHAKU FLIPPING (1–3 SWING-UP, 4–6 CHAIN ROLLING, 7–8 RE-GRASPING)

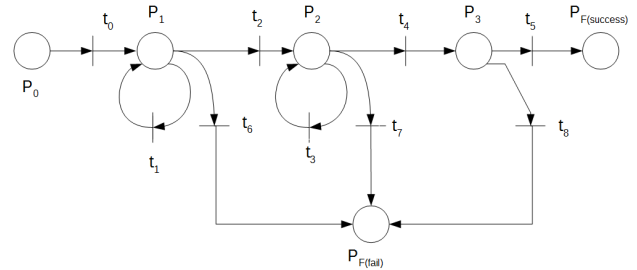

**FIGURE 10.** HUMAN DEFINITION OF THE FLIPPING SKILL USING A PETRI NET

Using the initial definition, the robot learns the control law of each transition specified in the Petri net by processing the recorded human demonstration (Fig. 9) with nonparametric learning. Sensing the motion variables of the nunchaku directly requires advanced and often unavailable sensing systems. As a workaround, the motion of the rods and the chain is estimated in real time by monitoring the centrifugal load on the end-effector. After a robot trial is completed, the final score and the score of every sub-procedure are obtained from applying an SVM for grading. If the trials constantly fall into the failure cluster, the robot modifies the initial definition by identifying the problematic part in the definition and make adjustment using the proposed APN. Figure 11 shows a learned successful performance (grasping is simplified as controlled magnetic attraction).

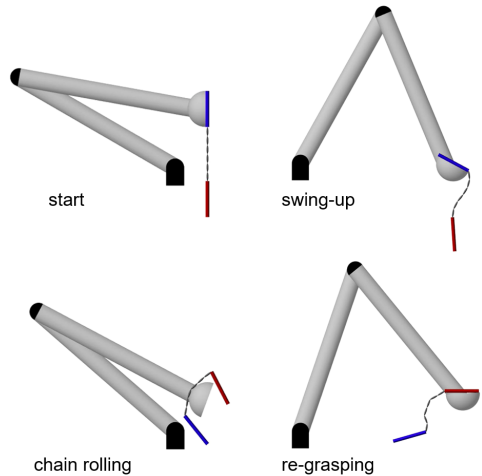

**FIGURE 11**. NUNCHAKU FLIPPING LEARNED BY A ROBOT IN SIMULATION

## CONCLUSIONS AND FUTURE WORK

This paper introduced a multiform robot learning scheme. The approach extends the single form of robot learning from human demonstration by introducing learning from abstract definition and autonomous evaluation. Together, these complementary forms of robot learning allow humans to teach robots in an intuitive manner that is similar to teaching humans, where lecturing (of definitions), imitations (to demonstrations), and grading (of practicing trials) are integrated in school settings. The work aims for enabling robots to efficiently obtain advanced skills that have been difficult for them to learn thus far, especially those requiring high dynamics, sophisticated contact control, accurate timing, and handling partly rigid partly soft gadgets. Adaptive approaches have been proposed to allow the robot to autonomously perceive human intentions in teaching through the demonstration data and evaluation.

Two tests have been conducted to validate the proposed robot learning scheme, including teaching an actual robot to swing up and balance an inverted pendulum, and teaching a simulated robot to flip a nunchaku, both showed satisfying effectiveness. Future work will focus on three parts: (1) conducting the second (the simulated) test on actual robots with kinesthetic teaching, (2) introducing fusion techniques to allow the robot to learn from multiple mentors, and (3) conducting further tests on more types of objects and skills.